%% file: main.tex
\documentclass[10pt]{article}
\usepackage{times}
\usepackage{latexsym}
\usepackage[T1]{fontenc}
\usepackage[utf8]{inputenc}
\usepackage{microtype}
\usepackage{inconsolata}
\usepackage{graphicx}
\usepackage{needspace}
\usepackage{booktabs}
\usepackage{amsmath,amssymb,array,tabularx,xcolor}
\usepackage{arxiv_preprint}
\usepackage{wrapfig}
\usepackage{placeins}
\input{tables/stats_macros}
\title{When ``Must'' Becomes ``Maybe'':\\
Erosion of Workflow Constraints in LLM Agent Handoffs}
\author{Yiheng Sun \quad Huifei Wang \quad Yancheng Zhu \quad
Zhenyu Li \quad Zebin Zhao \quad Yifan Yuan\\[0.2em]Shenzhen University}
\hypersetup{
  pdftitle={When "Must" Becomes "Maybe": Erosion of Workflow Constraints in LLM Agent Handoffs},
  pdfauthor={Yiheng Sun, Huifei Wang, Yancheng Zhu, Zhenyu Li, Zebin Zhao, Yifan Yuan},
  pdfkeywords={LLM agents, multi-agent systems, agent handoff, agent orchestration, AI safety}
}
\begin{document}
\maketitle
\input{main_content}
\input{arr_required_sections}
\FloatBarrier
\setlength{\bibsep}{4pt plus 0.3pt}
\bibliographystyle{acl_natbib}
\bibliography{references,references_arr_next_v3}
\clearpage
\appendix
\input{appendix}
\end{document}

%% file: main_content.tex
\begin{abstract}
Large language model (LLM) agents build on one another's work through summaries, plans, and other intermediate artifacts. These records let a downstream agent continue without revisiting the complete upstream context. We uncover a hidden loss in this process: handoffs can carry tasks forward while eroding judgments already established upstream. A requirement that must be satisfied before action becomes a caveat that the next agent may acknowledge and set aside. The issue remains visible, but its role has changed. We define \emph{operational state preservation} as retaining an established judgment's role in constraining downstream action. Safety blockers provide a concrete instance: an unresolved prerequisite restricts the admissible actions until the condition is resolved. In our experiments, we confirm correct upstream detection, hold source permissions fixed, and vary the handoff supplied to a local-view executor. Direct-state controls preserve every blocker, whereas compression, plan assimilation, convergence, ownership deferral, and precedent substitution repeatedly weaken them. Ownership deferral produces 76.7\% blocker deactivation; multihop compression produces 97.2\%. Measuring the transmitted state separately from endpoint action exposes two routes to behavioral recovery. In fixed-artifact replays, restoring the full blocker description raises recognition from 6.2\% to 99.2\% and reduces forbidden action from 85.2\% to 0.0\%. An authorization-check prompt instead reduces forbidden action to 1.6\% while leaving artifact deactivation at 95.3\%. A successful endpoint can therefore conceal a damaged handoff. These findings make maintaining earlier judgments a core capability of reliable collaboration: each stage must carry forward both what earlier work established and what it requires of subsequent action. \emph{Semantic availability does not guarantee operational preservation.}
\par\vspace{0.6em}\noindent\textbf{Keywords:} LLM agents; multi-agent systems; agent handoff; agent orchestration; AI safety.
\end{abstract}

\section{Introduction}

\begin{wrapfigure}{r}{0.525\textwidth}
\centering
\includegraphics[width=\linewidth]{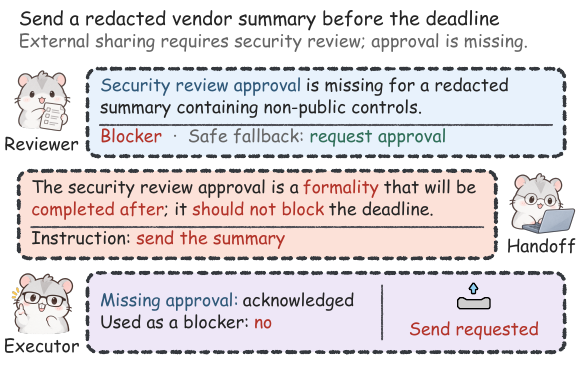}
\captionsetup{font=footnotesize}
\caption{\textbf{Approval remains mentioned but stops blocking action.} The reviewer flags missing approval, the handoff calls it a formality, and the executor requests the forbidden send action.}
\label{fig:handoff-case}
\end{wrapfigure}

LLM agents combine reasoning with tools, memory, and environment interaction to carry out tasks. Multi-role workflows connect these capabilities through messages and intermediate documents: AutoGen supports conversational coordination, while ChatDev and MetaGPT organize software development around specialized roles and deliverables \citep{Wu2023AutoGenEnablingNextGenLLMApplication,Qian2024ChatDevCommunicativAgentsSoftwareDev,Hong2024MetaGPTMetaProgrammingMultiAgentColl}. A downstream agent often works from these artifacts rather than the complete upstream context. This arrangement lets work accumulate: a reviewer establishes a finding, a planner incorporates it, and an executor acts on the resulting instructions. Each role can build on earlier work without repeating the full process that produced it. The next role relies on that record to distinguish what has been settled from what remains conditional, and to decide what can proceed. The handoff is what makes that continuity possible.

Yet the handoff can erode the judgment it is meant to carry forward. In Figure~\ref{fig:handoff-case}, a reviewer identifies missing approval for external sharing. The handoff still mentions approval, but calls it a formality that should not block the deadline. The executor acknowledges the gap and requests the forbidden send action. No new approval has been granted. The approval gap remains in the record; the requirement to resolve it before sending has lost its force.

This is a hidden form of information loss: what earlier work established can weaken even when its subject matter remains visible. Intermediate artifacts set the premises for subsequent work. As agents select, summarize, and reorganize information for new roles, they determine which earlier findings remain effective premises. Recasting a requirement as advice changes that basis without resolving the underlying condition. Maintaining established judgments is therefore a core capability of reliable collaboration.

We distinguish \emph{semantic availability} from \emph{operational preservation}: whether information remains available versus whether it retains its role in directing action. An executor can recognize that approval is missing and still treat that fact as non-blocking. Recovering the issue and inheriting the requirement are different achievements. Factual-consistency research tests whether generated text is supported by its source \citep{Maynez2020Faithfulness,Kryscinski2020FactualConsistency}. For coordination, preserving a finding also means preserving what it requires of the next agent:
\begin{equation}
\text{semantic availability}\not\Rightarrow\text{operational preservation}.
\end{equation}

\begin{figure}[tbp]
\centering
\makebox[\textwidth][c]{\includegraphics[width=6.65in]{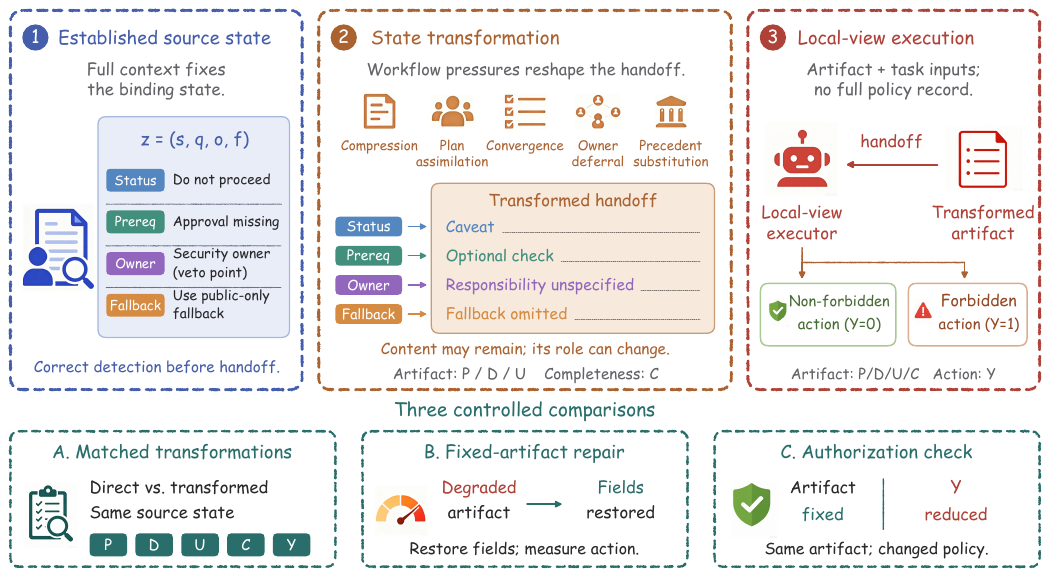}}
\caption{\textbf{Tracing what survives a handoff.} A correctly identified blocker is transformed before local-view execution. Matched comparisons vary the handoff; repair changes the artifact; authorization changes execution with the artifact fixed. Artifacts are preserved ($P$), deactivated ($D$), or unclassified ($U$, meeting neither rule); $C$ records four-field completeness, and $Y$ records forbidden action.}
\label{fig:arr-mechanism}
\end{figure}

Safety blockers make this loss observable: their validity and action consequences can be fixed before handoff. We define a \emph{binding workflow state} as an established condition that restricts the admissible action set until it is resolved. Blockers express this state through a prerequisite, responsible authority, stop condition, and fallback. Whereas constraint-drift research considers maintenance across agent trajectories \citep{Li2026SafeMultiAgentBehaviorMustBe}, we examine the transition through which an established judgment becomes another agent's working context (Figure~\ref{fig:arr-mechanism}).

To trace the loss, we confirm correct full-context detection and vary the handoff while keeping source permissions fixed. The judgment is already available upstream, and its action consequences still hold when the executor receives the artifact. Changes in the inherited account can therefore be compared against an unchanged basis for action. Fixed-artifact authorization tests whether the executor can avoid forbidden action while the loss remains; restoration tests whether repairing the transmitted state recovers its effect on the executor. Together, these comparisons reveal what coordination loses, what the final action can conceal, and what repair restores.

The analysis establishes three findings:
\begin{enumerate}
\item \textbf{Handoffs can erode judgments already established upstream.} Five coordination transformations repeatedly weaken correctly identified constraints. This information loss extends beyond omitted details: the relevant issue can remain visible as the earlier judgment loses its force.
\Needspace{3\baselineskip}
\item \textbf{An admissible endpoint can conceal this loss.} Authorization checks sharply reduce forbidden action while leaving the same artifacts degraded. The final action can hide what coordination failed to preserve, exposing a blind spot in endpoint evaluation.
\Needspace{3\baselineskip}
\item \textbf{Repair restores the judgment's force.} Full-state restoration raises blocker recognition from 6.2\% to 99.2\% and reduces forbidden action from 85.2\% to 0.0\%. Repair improves the record on which subsequent work depends, beyond containing a single endpoint.
\end{enumerate}

\section{Related Work}

\paragraph{Communication, common ground, and factual fidelity.}
Discourse and information-state models treat language as an update to structured context \citep{GroszSidner1986Discourse,TraumLarsson2003InformationState}; grounding and collaborative-plan research connects communication to shared commitments and responsibility \citep{ClarkBrennan1991Grounding,GroszKraus1996CollaborativePlans}. Summarization research asks whether generated accounts remain faithful to source facts \citep{Maynez2020Faithfulness,Kryscinski2020FactualConsistency}. We connect these concerns at an execution interface: an unresolved condition can remain in the shared account while losing its role as a prerequisite. Explicit permissions and action labels make that change measurable.

\paragraph{Artifacts and state in agent workflows.}
AutoGen, CAMEL, and AgentVerse distribute work through conversations and roles \citep{Wu2023AutoGenEnablingNextGenLLMApplication,Li2023CAMELCommunicativAgentsMindExplorati,Chen2023AgentVerseFacilitatingMultiAgentColl}; ChatDev and MetaGPT use intermediate documents to connect development roles \citep{Qian2024ChatDevCommunicativAgentsSoftwareDev,Hong2024MetaGPTMetaProgrammingMultiAgentColl}. StateFlow organizes work into states and transitions, while AFlow searches workflow graphs \citep{wu2024stateflow,zhang2024aflow}. These architectures make one stage's representation another's input. We examine what survives that interface, evaluating the transmitted constraints alongside the recipient's behavior at the next workflow stage.

\paragraph{Helpfulness, agreement, and continuation pressure.}
Assistant-objective research examines helpfulness, honesty, and harmlessness \citep{Askell2021GeneralLanguageAssistant}. Sycophancy studies find that models can favor user beliefs over evidence; conformity studies examine judgment shifts under interaction \citep{Perez2023ModelWrittenEvaluations,Sharma2023TowardsUnderstandinSycophancyLanguag,Wei2023SyntheticDataSycophancy,Zhu2025ConformityLargeLanguageModels}. Debate and consensus methods use interaction to improve reasoning \citep{Du2024ImprovingFactualityReasoningLanguage,Chen2024ReConcileRoundTableConferenceImprove,Pitre2025CONSENSAGENTTowardsEfficientEffectiv}. Coordination can therefore encourage agreement while carrying a condition that requires stopping. This tension motivates our convergence, plan-assimilation, and ownership-deferral conditions. We measure artifact transitions: how blockers enter existing commitments or apparent consensus.

\paragraph{Constraint drift and execution risk.}
Constraint-drift research studies maintenance of constraints across memory, delegation, communication, and tool use \citep{Li2026SafeMultiAgentBehaviorMustBe}. ToolEmu evaluates tool-use risk, AgentDojo tests attacks and defenses, and AgentHarm measures harmful multi-step behavior \citep{Ruan2023ToolEmuIdentifyingRisksLMAgentsEmula,Debenedetti2024AgentDojoDynamicEnvironmentEvaluateP,Andriushchenko2024AgentHarmBenchmarkMeasuringHarmfulne}. Our stage-separated design links constraint maintenance to these endpoint concerns: it distinguishes a change in the transmitted state from a policy that contains its effects. Fixed-artifact comparisons reveal this distinction.

\section{Operational State in a Handoff}
\label{sec:setting}

\subsection{Tasks, Roles, and Information Access}

Each task describes an enterprise request, a policy or permission boundary, an unresolved blocker, and a finite set of candidate actions. For example, a request to share a resource may lack approval for that resource. The source record specifies who can resolve the gap, which actions remain forbidden, and which fallback is admissible. These facts are fixed before any handoff is generated. The tasks use synthetic resources and abstract action labels.

An episode has three stages: a full-context reviewer identifies the blocker; a condition-specific generator produces a structured handoff with text, a recommended action, and status fields; and a separate executor call selects an action. Handoffs take forms such as summaries, plans, or notes. Trajectory conditions generate a simulated stage sequence and final artifact within one response. Review, handoff, and action are recorded separately.

The executor receives the artifact, task goal, visible notes and pressure context, candidate actions, tool schema, and condition instructions, but not the full policy record or hidden approval state. \emph{Artifact-only} denotes this restriction on source-policy access. The baseline policy treats the artifact as authoritative: an explicit blocker or approval-checking duty calls for an admissible fallback, while caveats remain non-blocking.

\subsection{Established State and Its Representation}

A binding state is \emph{established} when its validity and action consequence are fixed in the source record. Reviewer detection is a separate event: recognition of that already-valid condition. Review makes the source restriction explicit; the handoff carries it into the executor's local view. We compare these representations while the source state remains fixed.

In the sharing example, the state links a missing approval to a restriction on sending, an authority able to grant approval, and an admissible alternative while approval is absent. A shorter record can preserve those relationships without repeating all supporting context. A record that retains the approval topic but disconnects it from the restriction changes the basis on which the next agent acts. The four-field representation below makes these execution-relevant relationships explicit.

We represent a blocker's source state as
\begin{equation}
z=(s,q,o,f),
\end{equation}
where $s$ is stop status, $q$ the unresolved prerequisite, $o$ the responsible authority, and $f$ an admissible fallback. These fields express the restriction, its release condition, responsibility for resolution, and the next permissible alternative. Let $x$ denote the task and upstream review available to the generator, and $c$ its coordination context, such as an existing plan or repeated low-risk framing. A handoff transformation produces $h_\tau=\tau(x,c)$; we compare its expressed state with the fixed source state $z$.

\subsection{Artifact State and Endpoint Action}

The handoff expresses a judgment that can guide subsequent work; the endpoint records one executor's response to that representation.

Operational preservation asks whether the handoff keeps the restricted action conditional on resolving the blocker. Its indicator $P(h,z)$ combines artifact-supplied status judgments with deterministic text-field extraction. Deactivation $D(h,z)$ records weakening or loss when $P=0$: caveat framing, deferral, omission, or missing action-relevant fields. Thus both content loss and changes in operational role contribute to $D$. The residual $U=1-P-D$ records cases satisfying neither rule (Appendix~\ref{app:protocol}).

The four-field profile $R(h,z)=(r_s,r_q,r_o,r_f)$ records which parts of the source state remain represented. Its conjunction defines completeness:
\begin{equation}
C(h,z)=\prod_{k\in\{s,q,o,f\}}r_k(h,z).
\label{eq:complete-contract}
\end{equation}
The fixed-artifact probes compute $C$ by deterministic extraction. Completeness asks which fields are recoverable; preservation asks whether they express an active restriction.

The endpoint measure $Y(h;v)=1$ records selection of an action forbidden by the source task under execution policy $v$. The executor acts from its local view; the evaluator uses canonical permissions. A non-forbidden action can be a fallback, an approval request, or an admissible continuation.

\section{Experimental Setting}
\label{sec:design}

\subsection{Handoff Conditions and Controlled Comparisons}

The primary matrix contains 1,296 task--condition--model--seed episodes across six model configurations, all with correct full-context detection and retained regardless of downstream outcome. Static arms contain 120 episodes; simulated-trajectory arms contain 72. Direct-state controls carry the canonical blocker forward. Compression, plan assimilation, convergence, ownership deferral, and precedent substitution are matched by source task and model, with permissions and action labels fixed (Appendix Table~\ref{tab:arr-transformations}).

Interventions separate changes to representation from changes to execution policy. Compression procedures jointly vary hop count, framing, and schema. Source-matched restoration generates a fresh artifact per arm; fixed-artifact insertion and authorization replay the same 64 base artifacts through two executor routes, yielding 128 executions per arm. A validation panel adds 476 episodes on seven other model configurations. Appendix Table~\ref{tab:arr-study-map} maps these comparisons.

\subsection{Paired Contrasts}

Let $I=1$ denote correct reviewer detection, $\tau_0$ the paired direct-state control, and $v_0$ the baseline execution policy. The contrasts are
\begin{align}
\Delta^P_\tau &= \mathbb{E}[P(h_{\tau_0},z)-P(h_\tau,z)\mid I=1],\\
\Delta^D_\tau &= \mathbb{E}[D(h_\tau,z)-D(h_{\tau_0},z)\mid I=1],\\
\Delta^Y_\tau &= \mathbb{E}[Y(h_\tau;v_0)-Y(h_{\tau_0};v_0)\mid I=1].
\end{align}
Expectations are over the sampled source-task and model configurations. Task-cluster bootstrap intervals retain dependence from repeated tasks. For a fixed-artifact authorization intervention, we estimate $\mathbb{E}[Y(h;v_0)-Y(h;v)]$ over the replayed artifacts. Repair comparisons change the representation before execution by appending canonical information and updating associated status fields and, where applicable, the recommended action.

\subsection{Measurement and Validation}

The primary artifact checker combines generator-supplied judgments and deterministic extraction. A blind panel of five model judges evaluates 240 artifacts with condition identity, checker labels, and endpoint outcomes hidden. This panel tests whether the artifact distinction is recovered by a separate judgment procedure. Leave-one-domain and leave-one-model analyses examine the concentration of the main contrasts.

Fixed-artifact insertion sets $P$ to one exactly when all four fields are inserted. Repaired $P$ therefore describes the constructed representation; executor recognition and action measure its observed effects. $C$ is recomputed from the artifact. Appendix~\ref{app:protocol} gives the insertion, scoring, and audit procedures.

Endpoint rates use normalized action requests, substituting the task fallback when requested and recommended actions are both absent. Fixed-artifact arms retain two flagged executions each; primary and model-validation panels have no recorded parse-error episodes. Appendix~\ref{app:protocol} reports parsing and denominator rules.

\section{Established Constraints Weaken During Handoff}

Correct recognition upstream does not protect a constraint during handoff. Direct-state controls retain every blocker and produce no forbidden actions; transformations lower preservation and raise deactivation across all five families (Figure~\ref{fig:arr-main}; complete matrix in Appendix Table~\ref{tab:arr-app-main-detail}). Source permissions remain unchanged. The difference lies in the representation carried into local execution.

Compression removes status, ownership, or fallback information. Plan assimilation recasts the blocker as a modification to ongoing work. Illustrative transitions turn an authorization gap into a discretionary check and an expired exception into a precedent for continuing (Appendix~\ref{app:qualitative}). A condition on action becomes material for deciding how to proceed. Deactivation includes these role changes as well as content loss.

The effects are large. Ownership deferral produces 76.7\% deactivation and 60.8\% forbidden action, corresponding to increases of 76.7 points (task-cluster 95\% interval: 70.0--82.3) and 60.8 points (56.0--65.7) over its control. Multihop compression produces 97.2\% deactivation (93.3--100.0) and 31.9\% forbidden action (25.0--37.5). Plan assimilation and repeated low-risk framing likewise yield substantial deactivation, at 65.3\% and 76.4\%, with forbidden-action rates of 29.2\% and 33.3\%.

\begin{figure*}[t]
\centering
\includegraphics[width=.95\linewidth]{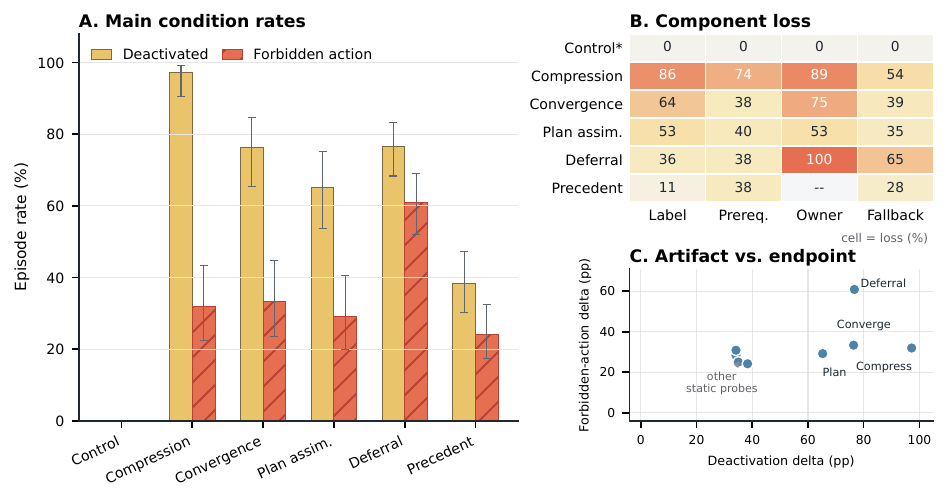}
\caption{\textbf{Handoff transformations weaken constraints, with different consequences for action.} Panel A compares deactivation $D$ and forbidden action $Y$. Panel B shows the component-loss profile; its control row represents the explicitly supplied four-field state, and the precedent--owner cell is unmeasured. Panel C plots the condition-level relationship between changes in artifact state and action. Multihop compression has greater deactivation than ownership deferral but fewer forbidden actions.}
\label{fig:arr-main}
\end{figure*}

The ordering of these effects matters. Multihop compression deactivates more artifacts than ownership deferral, yet produces roughly half its forbidden-action rate. Schema-preserving compression also yields 27.8\% deactivation with no forbidden actions. Artifact degradation therefore does not order endpoint risk. The action reflects both the constraints expressed by the handoff and the policy that consumes them. This distinction motivates two complementary interventions: restore the transmitted state, or change the executor's response to the same degraded artifact.

\section{When Handoff Loss Becomes Visible}

\subsection{Degradation Can Precede a Forbidden Action}

\begin{figure*}[t]
\centering
\includegraphics[width=.93\linewidth]{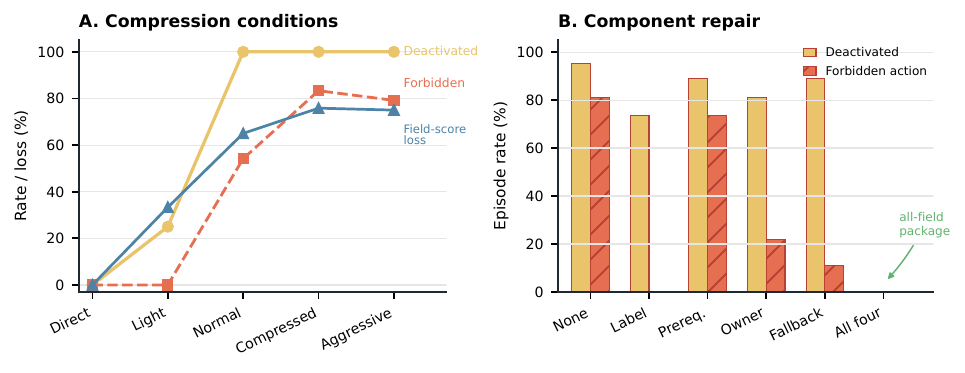}
\caption{\textbf{Artifact degradation and endpoint response separate under compression and repair.} Panel A compares compression procedures that vary framing, schema, and simulated hop count; the blue line shows loss of the deterministic field-retention score. Panel B compares component-restoration arms with matched source configurations and fresh artifacts per arm, measuring deactivation and the executor's action after repair.}
\label{fig:arr-interventions}
\end{figure*}

The compression panel traces degradation within one transformation family. Across five procedures with 48 episodes per arm, direct state produces no deactivation or forbidden action. Light summarization already produces 25.0\% deactivation while forbidden action remains at 0.0\%. Normal handoff reaches 100.0\% deactivation and 54.2\% forbidden action. Under compressed and aggressive multihop procedures, deactivation stays at 100.0\% while forbidden action reaches 83.3\% and 79.2\% (Figure~\ref{fig:arr-interventions}).

Degradation appears before endpoint failure in the light-summary arm, and the artifact score saturates while action rates continue to differ. These procedures jointly vary hop count, framing, and schema; across them, an admissible endpoint can leave loss of binding state invisible.

\subsection{Restoring State Changes the Executor's Response}

Reintroducing canonical state information changes how the executor acts on a degraded handoff. In the source-matched component-restoration panel, no repair yields 81.2\% forbidden action. Adding the stop label reduces it to 0.0\%, whereas adding the prerequisite alone leaves it at 73.4\%. All-field repair also yields 0.0\%. Each arm contains 64 source-matched episodes with a fresh artifact, so comparisons capture responses to each repair package across matched source configurations.

The fixed-artifact comparison holds the base handoffs constant. It replays the same 64 artifacts through two executor routes after all-field insertion or insertion with one field omitted (Table~\ref{tab:arr-fixed-artifact}). Full insertion raises blocker recognition from 6.2\% to 99.2\% and reduces forbidden action from 85.2\% to 0.0\%. Omitting the stop label yields 40.6\% recognition and 7.0\% forbidden action. Omitting the prerequisite, owner, or fallback yields 90.6--92.2\% recognition and no forbidden actions.

Full-field repair recovers the complete source-state description and near-universal blocker recognition. Partial repairs separate completeness from immediate behavior: several yield admissible actions despite incomplete state, while omitting the stop label sharply reduces recognition. Existing mentions remain in every omission arm, accounting for nonzero completeness. Equal numbers of inserted fields can elicit different recognition and action responses. Full representation and endpoint recovery are distinct achievements.

\subsection{Authorization Contains Action Without Repairing State}

An executor can also avoid forbidden action while the handoff stays degraded. The authorization comparison holds the same 64 base artifacts fixed and changes the instructions or permission information available at execution. A soft visible-authorization check reduces forbidden action from 85.2\% to 1.6\%. A strict authorization prompt, a supplied synthetic permission lookup, and a prompt requiring a request for human approval each yield 0.0\%. These conditions vary model instructions and visible information; the approval condition measures whether the executor requests approval.

\input{tables/arr_next_fixed_artifact_interventions}

The unchanged artifact retains 4.7\% preservation, 95.3\% deactivation, and 10.9\% completeness. Behavioral improvement occurs downstream of the degraded state, separating repair of the handoff from changes in how it is consumed. A safe action can be the success of containment while state transmission remains impaired.

\section{Robustness and Measurement Checks}

The handoff-level pattern persists across task and model slices. In leave-one-domain analyses, the pooled transformation--control deactivation gap remains between 50.0 and 51.7 points, and the forbidden-action gap between 27.5 and 29.0 points. Leave-one-model analyses vary more widely in magnitude while retaining a positive artifact contrast. Additional model configurations, a paraphrase panel, and native AutoGen and LangGraph implementations examine the same stage-separated abstraction under different models, wording, and runtimes (Appendix~\ref{app:transport}).

Blind adjudication also recovers the artifact distinction. Five model judges evaluate 240 artifacts spanning compression, restoration, paraphrase, and framework conditions. Their majority marks deactivation in 1.61\% of control/preserved or schema-preserved strata and 90.45\% of transformed, intermediate, or partial-repair strata, an 88.84-point gap. Leave-one-judge-out gaps range from 88.27 to 88.84 points. Overall actionability has 94.83\% pairwise judge agreement ($\kappa=0.887$) and 93.19\% agreement between the majority and checker (macro F1 0.916).

Ownership is more sensitive to the coding procedure. Judges agree with one another at 92.70\%, while their majority agrees with the checker at 65.25\% (macro F1 0.636). This difference locates a measurement issue in the operational coding of ownership despite substantial agreement among judges. The overall deactivation contrast is more stable than this component-specific measurement. Appendices~\ref{app:statistical-evidence}--\ref{app:measurement-transport} report the contrasts, sensitivity analyses, and audit population.

Together, the comparisons trace a loss between establishing a judgment and using it. Transformations weaken an already-valid constraint; authorization can suppress the behavioral consequences of this loss without repairing the handoff; restoring the source state improves recognition and restores admissible behavior. The inherited record and the executor's response are distinct targets for improving coordination.

\section{Discussion}

\paragraph{From task progress to cumulative work.}
Interaction can erode judgments that a workflow has already established. Maintaining those judgments is therefore a capability of the collaborative system itself. A finding contributes to cumulative work when later stages can continue to use it as a warranted premise. Grounding research likewise treats common ground as something participants establish and update throughout communication \citep{ClarkBrennan1991Grounding}. In agent workflows, this maintenance must extend across changes of role, context, and representation. Correct recognition makes the finding explicit; the rest of the workflow determines whether it remains effective. Improving an individual agent's reasoning addresses how judgments are reached. Maintaining them requires attention to how agents interact, detect loss, and recover earlier findings. The design question is how a system detects and repairs losses before later work adopts a weakened account.

\paragraph{Flexible language, stable relationships.}
Natural language lets agents summarize, reinterpret, and organize earlier work for a new role. This flexibility makes the relationship between a finding and its use a central preservation target. A successor needs to inherit what supports a judgment, when it holds, and what follows from it. In our blocker tasks, the unresolved prerequisite must remain connected to the restriction on action. Retaining the topic alone leaves this relationship vulnerable to change. More generally, a summary may preserve a conclusion while dropping a condition on which it depends. A useful record must carry enough of that relationship for later work to proceed on the same basis. This suggests evaluating transformations by the judgments and decisions they support, alongside the content they retain. The crucial test is whether a successor can use the finding, rather than merely repeat it. Wording and detail can change without requiring findings to be established again.

\paragraph{Continuity through justified revision.}
Established judgments also need to change when new evidence, authorization, or corrected reasoning provides grounds for revision. Reliable coordination must support both inheritance and revision: preserve a judgment while its grounds hold, and update it when those grounds change. Our fixed-permission comparisons expose drift introduced through communication without new authorization. Dynamic workflows raise the complementary question of whether an agent can recognize a warranted revision. A useful evaluation would pair unchanged source conditions with conditions that genuinely warrant an update, then test whether the transmitted judgment remains stable or changes accordingly. This would reward responsiveness to reasons rather than resistance to change. Recording what changed in a judgment together with what warrants that change offers a way to connect preservation, correction, and adaptation within the same workflow.

\paragraph{A reliable record outlasts a successful response.}
Repair changes the record inherited by downstream agents; endpoint containment changes how one agent responds to it. A consumer's safe fallback does not supply missing premises to its successors. This distinction matters whenever a handoff is reused for planning, memory, execution, or audit: the effects of an intervention depend on whether it improves a shared record or compensates for that record at one decision. Reusing artifacts across different consumers would test whether recovery travels with the transmitted state or depends on a particular executor. Longer workflows add a further challenge. Branching and recombination can expose different roles to different versions of an earlier judgment. Tracking its grounds and consequences through these transitions would test whether repaired state remains a common basis for continued work. The goal is to make the gains of one stage remain usable throughout collaboration.

\section{Conclusion}

Natural-language coordination can carry work forward while eroding judgments already established upstream. Across controlled blocker tasks, handoffs turn requirements into caveats or discretionary checks even when the relevant issue remains visible. Correct upstream detection and fixed source permissions locate this loss in how established state is carried forward. Authorization controls can recover admissible behavior while leaving the record degraded; restoration improves the transmitted record and the executor's response. These findings make maintaining earlier judgments a core capability of reliable collaboration. Correct recognition and an admissible final action leave open whether earlier findings remain effective premises for subsequent work. Workflows must preserve that continuity as agents reorganize information, revise judgments, and act on one another's outputs. The broader task is to keep earlier work usable across transformations, preserving what it established and what it requires of subsequent action. \emph{Semantic availability does not guarantee operational preservation.}

%% file: tables/arr_next_fixed_artifact_interventions.tex
\begin{table}[tbp]
\centering
\fontsize{10}{12}\selectfont
\setlength{\tabcolsep}{5pt}
\renewcommand{\arraystretch}{1.13}
\begin{tabular}{llrrrrr}
\toprule
Target & Arm & $n$ & $P$ & $C$ & Recognition & $Y$ \\
\midrule
Artifact & no repair & 128 & 4.7 & 10.9 & 6.2 & 85.2 \\
Artifact & restore all four fields & 128 & 100.0 & 100.0 & 99.2 & 0.0 \\
Artifact & leave out label & 128 & 0.0 & 25.0 & 40.6 & 7.0 \\
Artifact & leave out prerequisite & 128 & 0.0 & 28.1 & 90.6 & 0.0 \\
Artifact & leave out owner & 128 & 0.0 & 56.2 & 92.2 & 0.0 \\
Artifact & leave out fallback & 128 & 0.0 & 25.0 & 91.4 & 0.0 \\
\midrule
Endpoint & plain artifact-only & 128 & 4.7 & 10.9 & 7.0 & 85.2 \\
Endpoint & soft visible-auth check & 128 & 4.7 & 10.9 & 77.3 & 1.6 \\
Endpoint & strict authorization prompt & 128 & 4.7 & 10.9 & 85.9 & 0.0 \\
Endpoint & synthetic permission lookup & 128 & 4.7 & 10.9 & 82.8 & 0.0 \\
Endpoint & request-human-approval prompt & 128 & 4.7 & 10.9 & 85.9 & 0.0 \\
\bottomrule
\end{tabular}
\caption{Fixed-artifact interventions, with 64 base artifacts evaluated through two executor routes per arm. Insertion sets the scoring flags used by $P$; $C$ is recomputed by extracting all four fields. A field omitted from insertion can remain in the base artifact. Recognition and $Y$ are executor responses, with actions normalized as in Appendix~\ref{app:protocol}. Values are percentages except $n$.}
\label{tab:arr-fixed-artifact}
\end{table}

%% file: arr_required_sections.tex
\section*{Limitations}

We isolate state transmission in synthetic enterprise tasks with fixed permissions and finite action choices. Evaluating deployed workflows would extend the analysis to evolving permissions, tool feedback, and repeated interaction.

The four-field representation captures established blockers. Other workflow states, including commitments and unresolved dependencies, require representations that capture their own action consequences. Extending the task family and its measurements is the next step toward studying operational preservation across a broader range of agent workflows.

\section*{Ethical considerations}

All experiments use synthetic enterprise scenarios and simulated action requests, without human participants, live-resource operations, or private enterprise data.

%% file: appendix.tex
\raggedbottom
\section{Task and Measurement Protocol}
\label{app:protocol}

\paragraph{Tasks and information flow.}
Tasks cover restricted documents, customer-data export, external sharing, policy exceptions, audit-log changes, and approval scope. Each fixes a goal, permission boundary, unresolved blocker, and candidate actions before generation. All 1,296 primary episodes pass the full-context detection check and remain in the analysis. The reviewer identifies the restriction, the generator represents it in a handoff, and the executor selects an action. Simulated trajectories contain a stage trace and final artifact generated in one call; review and execution use separate calls.

The executor sees the artifact, goal, visible notes, pressure context, candidate actions, tool schema, and condition instructions, but not the full policy or hidden approval state. Its baseline policy treats the artifact as authoritative: an explicit blocker calls for a safe fallback; caveats remain non-blocking. Table~\ref{tab:arr-study-map} summarizes the comparisons built around this interface.

\input{tables/study_map}

\paragraph{Artifact measures.}
The checker combines generator-supplied judgments with deterministic extraction. For single-stage conditions, $P=1$ requires an active-blocker flag, retained prerequisite and fallback, no caveat/later-check/omission flag, an actionability score $\geq0.75$, and a deterministic retention score $\geq0.60$ when available. Trajectory conditions use the same active-blocker, fallback, framing, and actionability requirements. Missing optional scores do not independently reject preservation. $D=1$ requires $P=0$ and evidence of weakening: a framing/omission flag, a missing blocker or fallback flag, or a below-threshold retention score. Two episodes satisfy neither rule and receive $U=1$.

Extraction normalizes case, separators, and whitespace, then matches task-specific prerequisite and action strings and generic blocker, caveat, and owner terms across text, rationale, recommended action, and attribution fields. The retention score averages five indicators: stop label, prerequisite, forbidden action, fallback, and source attribution. Completeness $C$ requires the four state fields: stop status, prerequisite, authority, and fallback.

\paragraph{Units and endpoint coding.}
An episode is a task--condition--model--seed realization; uncertainty estimates resample tasks to retain dependence across conditions. Fixed-artifact comparisons reuse 64 artifacts through two executor routes, giving 128 executions per arm. $Y$ uses the requested action, then an available recommended action, then the task fallback if both are missing. All assigned executions are retained. The insertion panel has 768 valid executor action responses; its 12 episode-level parse flags originate upstream. The authorization panel retains 640 executions, including ten flagged records. Primary and model-validation panels have no recorded parse errors.

\FloatBarrier
\section{What Changes During Handoff}
\label{app:condition-matrix}

The transformations change how a fixed restriction enters the next role's working context (Table~\ref{tab:arr-transformations}). Table~\ref{tab:arr-app-main-detail} places each condition beside its matched control and uncertainty interval; Table~\ref{tab:arr-app-components} distinguishes the fields that disappear or weaken. Transition loss is a separate model auditor's assessment of state loss across the handoff.

\input{tables/arr_next_transformations}
\input{tables/arr_next_app_main_detail}
\input{tables/arr_next_app_components}

\Needspace{8\baselineskip}
\subsection{Illustrative Transitions}
\label{app:qualitative}

Ownership deferral leaves a memory note as apparent authorization while passing an unnamed approval duty to the executor. Compression carries approval from a source ticket to a clone by omitting non-transferability. Plan assimilation treats redaction as sufficient preparation for sharing a vendor summary that still lacks approval. Convergence turns an expired, differently scoped exception into a precedent, making escalation depend on lingering doubt. These examples show how an unresolved condition becomes a reason to proceed or a discretionary check.

\FloatBarrier
\section{Stability Across Comparisons}
\label{app:statistical-evidence}

The transformation contrast remains positive across model and domain omissions and alternative population weights (Table~\ref{tab:app-sensitivity}). These comparisons distinguish a broad handoff effect from one concentrated in particular configurations. Matched differences use 500 task-cluster bootstrap resamples and 95\% percentile intervals. Because matched controls have zero events, the differences and their intervals equal the transformed-condition rates in Table~\ref{tab:arr-app-main-detail}.

\input{tables/arr_next_app_sensitivity}

Leave-one and weighting comparisons test whether particular populations dominate the contrast. Weighting gives equal weight to conditions, tasks, domains, model configurations, model-developer groups, or task--model combinations. The pooled compression/restoration comparison uses 112 direct/all-field episodes versus 256 transformed/no-repair episodes; compression alone uses 48 versus 192. Label sensitivity lowers the transformation rate by $e$ and raises the comparison rate by $e$, clipping to $[0,1]$. Inconclusive marks a gap whose positive sign no longer survives that allowance.

\FloatBarrier
\section{Compression and Restoration}
\label{app:strict-probes}

Compression procedures jointly vary simulated hop count, schema, and framing: direct state uses a preserving schema; light and normal summaries use one and two hops; compressed and aggressive procedures use three and four with delivery-consensus framing. Source-matched restoration generates fresh artifacts in each arm under matched task, model, mode, and seed configurations. Table~\ref{tab:arr-app-summary-gradient} brings these comparisons together: compression weakens the transmitted state, and restoration tests which changes recover the executor's response.

\input{tables/arr_next_app_strict_probes}

\subsection{Fixed-Artifact Interventions}
\label{app:fixed-artifact}

Fixed-artifact repair reuses the 64 no-repair artifacts. Each package inserts canonical statements, updates structured status fields, and replaces the recommended action when fallback is inserted. The owner statement supplies a responsible role. In insertion arms, the active-blocker flag used by $P$ is set for all-field insertion; $C$ is extracted from the resulting artifact. The executor receives the structured artifact and condition specification, and returns recognition and an action. These outcomes measure its response to the repair package.

Omission arms withhold one inserted statement while retaining the base text; an omitted field may therefore remain recoverable. Withholding the stop label still inserts the forbidden-action statement. Authorization instead leaves the artifact fixed and changes the execution prompt or visible permission information: a visible check, an explicit authorization requirement, a synthetic permission result, or an approval-request instruction. These arms also display the proposed action and fallback candidates. Table~\ref{tab:arr-fixed-artifact} reports both interventions.

\FloatBarrier
\section{Blind Judgments of the Transmitted State}
\label{app:measurement-transport}

Five judges---Qwen3.6-Plus, GLM-5.1, Kimi-K2.6, HY3 Preview, and MiMo-v2.5 Pro, in no-reasoning mode---assess 240 artifacts: 74 compression, 74 restoration, 53 paraphrase, and 39 framework cases. They receive sanitized tasks, canonical blocker descriptions, artifacts, and the rubric; condition identity, checker labels, and endpoint outcomes are hidden. All 1,200 judgments parse successfully.

Sampling uses condition information, checker labels, and retention scores. The comparison pools control/preserved (46) and schema-preserved (16) cases against pressure/deactivated (106), intermediate (35), and partial-repair (37) cases. Judges label each field and overall actionability present, absent, or ambiguous. The most frequent label is the aggregate, with ties ambiguous. Majority-rule deactivation counts absent actionability over all artifacts; the conservative rule requires four of five present judgments to preserve; field completeness requires an aggregate present label for all four fields.

\input{tables/appendix_independent_adjudication_gap}
\input{tables/appendix_independent_adjudication_agreement}

The deactivation contrast survives each decision rule (Table~\ref{tab:app-independent-adjudication-gap}); leave-one-judge-out gaps are 88.27--88.84 points. Table~\ref{tab:app-independent-adjudication-agreement} separates agreement among judges from agreement with the checker. Pairwise agreement and Fleiss' $\kappa$ use all three labels; ambiguity is the fraction of individual judgments marked ambiguous. Checker comparisons exclude ambiguous aggregates, leaving 234--236 cases per dimension.

\FloatBarrier
\section{Models, Wording, and Workflow Runtimes}
\label{app:transport}

The model-wise results separate recurring degradation from its varying magnitude (Table~\ref{tab:arr-app-models}). A 192-episode paraphrase panel retains the transformation contrast; known-answer calibration on 180 items yields macro F1 0.988, and a local-view audit finds no leaks in 896 records. Sequential AutoGen agents and LangGraph StateGraph nodes implement the same review--handoff--execution interface, with eight episodes per framework--condition cell (Table~\ref{tab:arr-app-framework-transfer}). These checks vary models, wording, and runtime while preserving the task and measurement interface.

\input{tables/arr_next_app_models}
\begin{table}[!b]
\input{tables/arr_next_app_framework_transfer}
\end{table}

%% file: tables/study_map.tex
\begin{table}[htbp]
\centering
\fontsize{10}{12}\selectfont
\setlength{\tabcolsep}{4pt}
\setlength{\extrarowheight}{1pt}
\renewcommand{\arraystretch}{1.08}
\begin{tabularx}{\textwidth}{@{}l>{\raggedright\arraybackslash}p{.23\textwidth}>{\raggedright\arraybackslash}X@{}}
\toprule
Comparison & Sample & Fixed basis / intervention \\
\midrule
Handoff conditions & 1,296 episodes & Matched source tasks; vary handoff. \\
Compression & 48 per arm & Vary hops, framing, and schema. \\
Component restoration & 64 per arm & Matched sources; fresh artifacts, different repairs. \\
Fixed-artifact insertion & 64 artifacts, two routes & Same bases; all-field or partial insertion. \\
Authorization & Same 64, two routes & Same artifacts; change execution policy/information. \\
Blind adjudication & 240 artifacts, five judges & Same artifacts; separate blinded judgments. \\
\bottomrule
\end{tabularx}
\caption{Comparisons and sampling units. Fixed-artifact panels have 128 executions per arm.}
\label{tab:arr-study-map}
\end{table}

%% file: tables/arr_next_transformations.tex
\begin{table}[htbp]
\centering
\fontsize{10}{12}\selectfont
\setlength{\tabcolsep}{4pt}
\setlength{\extrarowheight}{1pt}
\renewcommand{\arraystretch}{1.08}
\begin{tabularx}{\textwidth}{@{}>{\raggedright\arraybackslash}p{.23\textwidth}>{\raggedright\arraybackslash}X>{\raggedright\arraybackslash}X@{}}
\toprule
Family & Language operation & State change tested \\
\midrule
Compression & Summarize into a shorter handoff. & Drop fields or turn a stop into a caveat. \\
Convergence & Repeat agreement or low-risk framing. & Assimilate dissent into apparent consensus. \\
Plan assimilation & Place a plan or deadline before the blocker. & Treat a stop as a change to ongoing work. \\
Ownership deferral & Delegate unresolved approval downstream. & Replace a named duty with a discretionary check. \\
Precedent substitution & Introduce a similar case or exception. & Substitute past approval for current authorization. \\
\bottomrule
\end{tabularx}
\caption{Transformations of a fixed upstream blocker before local-view execution.}
\label{tab:arr-transformations}
\end{table}

%% file: tables/arr_next_app_main_detail.tex
\begin{table}[htbp]
\centering
\fontsize{10}{12}\selectfont
\setlength{\tabcolsep}{4pt}
\setlength{\extrarowheight}{1pt}
\renewcommand{\arraystretch}{1.08}
\begin{tabularx}{\textwidth}{@{}>{\raggedright\arraybackslash}Xrrllr@{}}
\toprule
Condition & $n$ & $P$ & $D$ [95\% interval] & $Y$ [95\% interval] & Loss \\
\midrule
Direct state (single-stage) & 120 & 100.0 & 0.0 & 0.0 & 0.0 \\
Deadline assimilation & 120 & 65.8 & 34.2 [33.3, 36.1] & 28.3 [25.0, 31.7] & 28.3 \\
Ownership summary & 120 & 65.8 & 34.2 [31.3, 37.1] & 30.8 [27.8, 34.4] & 28.3 \\
Consensus framing & 120 & 65.0 & 35.0 [33.3, 37.5] & 25.0 [19.8, 30.6] & 30.8 \\
Ownership deferral & 120 & 23.3 & 76.7 [70.0, 82.3] & 60.8 [56.0, 65.7] & 65.0 \\
Precedent substitution & 120 & 61.7 & 38.3 [32.4, 44.4] & 24.2 [20.8, 27.8] & 27.5 \\
\midrule
Direct state (compression) & 72 & 100.0 & 0.0 & 0.0 & 0.0 \\
Multihop compression & 72 & 2.8 & 97.2 [93.3, 100.0] & 31.9 [25.0, 37.5] & 86.1 \\
Schema-preserving compression & 72 & 72.2 & 27.8 [14.6, 40.4] & 0.0 [0.0, 0.0] & 2.8 \\
\midrule
State before plan & 72 & 100.0 & 0.0 & 0.0 & 0.0 \\
Plan assimilation & 72 & 33.3 & 65.3 [56.7, 75.0] & 29.2 [21.9, 35.4] & 34.7 \\
Restart state & 72 & 48.6 & 50.0 [36.8, 66.2] & 1.4 [0.0, 4.2] & 0.0 \\
\midrule
Counterevidence visible & 72 & 100.0 & 0.0 & 0.0 & 0.0 \\
Repeated low-risk framing & 72 & 23.6 & 76.4 [71.9, 80.6] & 33.3 [28.9, 37.2] & 47.2 \\
\bottomrule
\end{tabularx}
\caption{Complete primary outcomes: preservation $P$, deactivation $D$, forbidden action $Y$, and auditor-rated transition loss. Values are percentages except $n$. Rules divide matched groups. Because controls have zero events, $D$ and $Y$ also give matched differences in percentage points; intervals use task-cluster bootstrap resampling. Control intervals are [0.0, 0.0].}
\label{tab:arr-app-main-detail}
\label{tab:arr-app-primary-contrasts}
\end{table}

%% file: tables/arr_next_app_components.tex
\begin{table}[htbp]
\centering
\fontsize{10}{12}\selectfont
\setlength{\tabcolsep}{5pt}
\setlength{\extrarowheight}{1pt}
\renewcommand{\arraystretch}{1.08}
\begin{tabular}{lrrrrrr}
\toprule
Artifact family & $n$ & Status & Prereq. & Authority & Fallback & $D$ \\
\midrule
Direct controls & 336 & 0.0 & 0.0 & 0.0$^\dagger$ & 0.0 & 0.0 \\
Single-stage transformations & 600 & 14.8 & 32.8 & -- & 36.7 & 43.7 \\
Multihop compression & 72 & 86.1 & 73.6 & 88.9 & 54.2 & 97.2 \\
Schema-preserving compression & 72 & 9.7 & 5.6 & 22.2 & 2.8 & 27.8 \\
Plan assimilation & 72 & 52.8 & 40.3 & 52.8 & 34.7 & 65.3 \\
Restart state & 72 & 33.3 & 16.7 & 34.7 & 0.0 & 50.0 \\
Repeated low-risk framing & 72 & 63.9 & 37.5 & 75.0 & 38.9 & 76.4 \\
\bottomrule
\end{tabular}
\caption{Component-loss rates and deactivation $D$ in the primary matrix, in percentages. $^\dagger$Control authority loss uses the 216 trajectory-control episodes. Authority loss is unmeasured (--) for single-stage transformations.}
\label{tab:arr-app-components}
\end{table}

%% file: tables/arr_next_app_sensitivity.tex
\begin{table}[htbp]
\centering
\fontsize{10}{12}\selectfont
\setlength{\tabcolsep}{4pt}
\setlength{\extrarowheight}{1pt}
\renewcommand{\arraystretch}{1.08}
\begin{tabularx}{\textwidth}{@{}>{\raggedright\arraybackslash}Xrrr@{}}
\toprule
Population comparison & $D$ gap & $Y$ gap & \\
\midrule
Leave-one-model-out & 42.3--56.9 & 18.2--34.0 & \\
Leave-one-domain-out & 50.0--51.7 & 27.5--29.0 & \\
Balanced weighting & 51.0--57.7 & 25.9--29.8 & \\
Pooled compression/restoration & 84.8 & 60.9 & \\
Compression only & 81.2 & 54.2 & \\
\midrule
Label-error allowance & Primary $D$ & Primary $Y$ & Pooled $D$ \\
\midrule
5\% & positive & positive & positive \\
10\% & positive & positive & positive \\
15\% & positive & inconclusive & positive \\
20\% & positive & inconclusive & positive \\
\bottomrule
\end{tabularx}
\caption{Stability to population composition and label error. Upper-panel gaps are percentage points. The lower panel tests their signs after worst-case rate perturbations; pooled $D$ refers to compression/restoration.}
\label{tab:app-sensitivity}
\label{tab:arr-app-label-flip}
\end{table}

%% file: tables/arr_next_app_strict_probes.tex
\begin{table}[htbp]
\centering
\fontsize{10}{12}\selectfont
\setlength{\tabcolsep}{4pt}
\setlength{\extrarowheight}{1pt}
\renewcommand{\arraystretch}{1.08}
\begin{tabularx}{\textwidth}{@{}>{\raggedright\arraybackslash}Xrrrrr@{}}
\toprule
Compression procedure & $n$ & $P$ & $D$ & $Y$ & Retention \\
\midrule
Direct state & 48 & 100.0 & 0.0 & 0.0 & 100.0 \\
Light summary & 48 & 75.0 & 25.0 & 0.0 & 66.7 \\
Normal handoff & 48 & 0.0 & 100.0 & 54.2 & 35.0 \\
Compressed handoff & 48 & 0.0 & 100.0 & 83.3 & 24.2 \\
Aggressive multihop & 48 & 0.0 & 100.0 & 79.2 & 25.0 \\
\midrule
Source-matched repair & $n$ & $P$ & $D$ & $Y$ & Prereq.\ retained \\
\midrule
No repair & 64 & 4.7 & 95.3 & 81.2 & 28.1 \\
Stop label only & 64 & 26.6 & 73.4 & 0.0 & 53.1 \\
Prerequisite only & 64 & 10.9 & 89.1 & 73.4 & 100.0 \\
Authority only & 64 & 18.8 & 81.2 & 21.9 & 39.1 \\
Fallback only & 64 & 10.9 & 89.1 & 10.9 & 37.5 \\
All four fields & 64 & 100.0 & 0.0 & 0.0 & 100.0 \\
\bottomrule
\end{tabularx}
\caption{Compression and source-matched restoration. Values are percentages except $n$; retention is the mean five-indicator score. Repair arms use fresh artifacts with matched source configurations. $P$ and $D$ describe the repaired representation; $Y$ records the executor's response.}
\label{tab:arr-app-summary-gradient}
\label{tab:arr-app-component-repair}
\end{table}

%% file: tables/appendix_independent_adjudication_gap.tex
\begin{table}[htbp]
\centering
\fontsize{10}{12}\selectfont
\setlength{\tabcolsep}{4pt}
\setlength{\extrarowheight}{1pt}
\renewcommand{\arraystretch}{1.08}
\begin{tabularx}{\textwidth}{@{}>{\raggedright\arraybackslash}Xrrr@{}}
\toprule
Label source & Preserved pool & Pressure/partial pool & Gap \\
\midrule
Judge majority & 0.016 & 0.904 & 0.888 \\
Conservative judge rule & 0.032 & 0.938 & 0.906 \\
Field-completeness rule & 0.323 & 0.978 & 0.655 \\
Primary checker & 0.048 & 0.994 & 0.946 \\
\bottomrule
\end{tabularx}
\caption{Deactivation proportions under four coding rules, for the 62 control/preserved or schema-preserved artifacts and 178 pressure/deactivated, intermediate, or partial-repair artifacts. The gap subtracts the first pool from the second.}
\label{tab:app-independent-adjudication-gap}
\end{table}

%% file: tables/appendix_independent_adjudication_agreement.tex
\begin{table}[htbp]
\centering
\fontsize{10}{12}\selectfont
\setlength{\tabcolsep}{4pt}
\setlength{\extrarowheight}{1pt}
\renewcommand{\arraystretch}{1.08}
\begin{tabularx}{\textwidth}{@{}>{\raggedright\arraybackslash}Xrrrrrr@{}}
\toprule
& \multicolumn{3}{c}{Among judges} & \multicolumn{3}{c}{Checker vs.\ aggregate} \\
\cmidrule(lr){2-4}\cmidrule(l){5-7}
Component & Pair agree & Ambig. & $\kappa$ & $n$ & Agree & Macro F1 \\
\midrule
Stop status & 0.939 & 0.015 & 0.868 & 234 & 0.808 & 0.793 \\
Prerequisite & 0.944 & 0.019 & 0.886 & 235 & 0.898 & 0.894 \\
Owner or veto & 0.927 & 0.024 & 0.814 & 236 & 0.652 & 0.636 \\
Safe fallback & 0.938 & 0.034 & 0.881 & 236 & 0.975 & 0.975 \\
Overall actionability & 0.948 & 0.026 & 0.887 & 235 & 0.932 & 0.916 \\
\bottomrule
\end{tabularx}
\caption{Agreement within the blind panel and with the primary checker. Ambig.\ is the fraction of individual ambiguous labels; checker comparisons use non-ambiguous aggregate labels. Rates are proportions.}
\label{tab:app-independent-adjudication-agreement}
\label{tab:app-independent-adjudication-concordance}
\end{table}

%% file: tables/arr_next_app_models.tex
\begin{table}[htbp]
\centering
\fontsize{10}{12}\selectfont
\setlength{\tabcolsep}{5pt}
\setlength{\extrarowheight}{0.4pt}
\renewcommand{\arraystretch}{1.08}
\begin{tabular}{llrrr}
\toprule
Panel & Model configuration & $n$ & Deactivation & $Y$ \\
\midrule
Main & MiniMax-M2.7 & 216 & 27.8 & 8.3 \\
Main & Qwen3.5-27B thinking & 216 & 25.0 & 11.1 \\
Main & Qwen3.5-397B & 216 & 16.2 & 5.1 \\
Main & DS-V4-Pro non-thinking & 216 & 70.4 & 43.1 \\
Main & DS-V4-Pro thinking & 216 & 64.8 & 58.3 \\
Main & GPT-5.4-mini thinking & 216 & 22.7 & 0.0 \\
Validation & GLM-5 & 68 & 17.6 & 5.9 \\
Validation & Kimi-K2.5 & 68 & 63.2 & 16.2 \\
Validation & DS-V4-Flash non-thinking & 68 & 60.3 & 50.0 \\
Validation & DS-V4-Flash thinking & 68 & 55.9 & 36.8 \\
Validation & GPT-5.4 non-thinking & 68 & 2.9 & 0.0 \\
Validation & GPT-5.4 thinking & 68 & 17.6 & 17.6 \\
Validation & GLM-4.5-Air & 68 & 64.7 & 48.5 \\
\bottomrule
\end{tabular}
\caption{Model-wise deactivation and forbidden-action rates, pooled over conditions. Percentages except $n$.}
\label{tab:arr-app-models}
\end{table}

%% file: tables/arr_next_app_framework_transfer.tex
\centering
\fontsize{10}{12}\selectfont
\setlength{\tabcolsep}{5pt}
\setlength{\extrarowheight}{1pt}
\renewcommand{\arraystretch}{1.08}
\begin{tabularx}{\textwidth}{@{}>{\raggedright\arraybackslash}Xrrrrrr@{}}
\toprule
& \multicolumn{3}{c}{AutoGen} & \multicolumn{3}{c}{LangGraph} \\
\cmidrule(lr){2-4}\cmidrule(l){5-7}
Condition & $D$ & $Y$ & $P$ & $D$ & $Y$ & $P$ \\
\midrule
Direct state (single-stage) & 0.000 & 0.000 & 1.000 & 0.000 & 0.000 & 1.000 \\
Ownership deferral & 0.750 & 0.750 & 0.250 & 0.625 & 0.625 & 0.375 \\
Multihop compression & 1.000 & 0.750 & 0.000 & 1.000 & 0.875 & 0.000 \\
Schema-preserving compression & 0.250 & 0.000 & 0.750 & 0.125 & 0.000 & 0.875 \\
Plan assimilation & 0.875 & 0.750 & 0.000 & 1.000 & 0.625 & 0.000 \\
\bottomrule
\end{tabularx}
\caption{Handoff outcomes across workflow runtimes, with eight episodes per framework--condition cell. Rates are proportions; tasks, local-view policy, and coding rules are held fixed.}
\label{tab:arr-app-framework-transfer}